\documentclass[sigconf,nonacm]{acmart}
\usepackage{multirow}
\usepackage{enumitem}
\usepackage{amsmath}
\makeatletter
\@ifundefined{Bbbk}{}{\let\oldBbbk\Bbbk\let\Bbbk\relax}
\makeatother
\usepackage{amssymb}
\makeatletter
\@ifundefined{oldBbbk}{}{\let\Bbbk\oldBbbk}
\makeatother
\usepackage{algorithm}
\usepackage{algorithmic}
\usepackage{graphicx}
\usepackage{caption}
\usepackage{subcaption}
\usepackage{hyperref}
\makeatletter
\newcommand\blfootnote[1]{%
  \begingroup
  \renewcommand\thefootnote{}\footnote{#1}%
  \addtocounter{footnote}{-1}%
  \endgroup
}
\makeatletter
\begin{document}

\title{ABLE: Using Adversarial Pairs to Construct Local Models for Explaining Model Predictions}

\author{Krishna Khadka}
\email{krishna.khadka@mavs.uta.edu}
\affiliation{%
  \institution{University of Texas at Arlington}
  \city{Arlington}
  \state{TX}
  \country{USA}
}

\author{Sunny Shree}
\email{sunny.shree@mavs.uta.edu}
\affiliation{%
  \institution{University of Texas at Arlington}
  \city{Arlington}
  \state{TX}
  \country{USA}
}

\author{Pujan Budhathoki}
\email{pxb9189@mavs.uta.edu}
\affiliation{%
  \institution{University of Texas at Arlington}
  \city{Arlington}
  \state{TX}
  \country{USA}
}

\author{Yu Lei}
\email{ylei@uta.edu}
\affiliation{%
  \institution{University of Texas at Arlington}
  \city{Arlington}
  \state{TX}
  \country{USA}
}

\author{Raghu Kacker}
\email{raghu.kacker@nist.gov}
\affiliation{%
  \institution{National Institute of Standards and Technology}
  \city{Gaithersburg}
  \state{MD}
  \country{USA}
}

\author{D. Richard Kuhn}
\email{d.kuhn@nist.gov}
\affiliation{%
  \institution{National Institute of Standards and Technology}
  \city{Gaithersburg}
  \state{MD}
  \country{USA}
}

\begin{abstract}
 Machine learning models are increasingly used in critical applications but are mostly "black boxes" due to their lack of transparency. Local explanation approaches, such as LIME, address this issue by approximating the behavior of complex models near a test instance using simple, interpretable models. However, these approaches often suffer from instability and poor local fidelity. In this paper, we propose a novel approach called Adversarially Bracketed Local Explanation (ABLE) to address these limitations. Our approach first generates a set of neighborhood points near the test instance, $x_{\text {test }}$, by adding bounded Gaussian noise. For each neighborhood point $D$, we apply an adversarial attack to generate an adversarial point $A$ with minimal perturbation that results in a different label than $D$. A second adversarial attack is then performed on $A$ to generate a point $A^{\prime}$ that has the same label as $D$ (and thus different than A). The points $A$ and $A^{\prime}$ form an adversarial pair that brackets the local decision boundary for $x_{\text {test }}$. We then train a linear model on these adversarial pairs to approximate the local decision boundary. Experimental results on six UCI benchmark datasets across three deep neural network architectures demonstrate that our approach achieves higher stability and fidelity than the state-of-the-art.
\end{abstract}


\maketitle
\blfootnote{This is a preprint of the accepted manuscript for ACM SIGKDD Conference on Knowledge Discovery and Data Mining (KDD 2026).}

\section{Introduction}
Modern machine learning (ML) models such as deep neural networks have become increasingly accurate, finding wide adoption in real-world applications ranging from healthcare diagnostics to financial forecasting \cite{lecun2015deep}. Despite their predictive power, these models are frequently criticized as unexplainable because they do not transparently reveal their internal decision-making processes \cite{lipton2016mythos, rudin2019stop,arrieta2020explainable}. In many high-stakes domains—such as hiring, loan approvals, or medical diagnostics—simply achieving high predictive accuracy is not enough as it is important to understand why a particular decision was made. 

Global explanation, e.g., overall feature importance \cite{linardatos2020explainable, lipton2018mythos}, capture general trends in a model’s behavior, yet they fail to reveal instance-specific factors that drive individual predictions. Local explanations focus on a single instance $x_{test}$ whose prediction we want to understand \cite{baehrens2010explain,doshi2017towards}. One common approach to local explanations is to build a simple, interpretable  model, such as a linear or rule-based predictor, that approximates the behavior of the global or target classifier $f(\cdot)$ in a neighborhood around $x_{test}$. The local model's coefficients or rules then provide feature attributions or explanations. Among the most widely recognized approaches that build local models as explanations are Local Interpretable Model-Agnostic Explanations (LIME) \cite{ribeiro2016should} and its variants.

Despite its popularity due to simplicity and model-agnostic nature, LIME-based explanation suffers from \textbf{instability} and \textbf{poor fidelity}. Its reliance on random perturbations makes the generated explanations sensitive to minor variations (instability), and its focus on the test instance, in terms of biased sampling that favors points near the test instance, can miss the true decision boundary of the target model, leading to inaccurate local approximations (poor fidelity) \cite{tan2024glime, alvarez2018robustness, guidotti2018survey, slack2020fooling}.

Recent advances work on these issues. GLIME \cite{tan2024glime} enhances stability by integrating weights into the sampling distribution, improving convergence. CALIME \cite{cinquini2024causality} enforces causal relationships between features in synthetic sampling, and USLIME \cite{saadatfar2024us} uses uncertainty sampling to target low-confidence regions. However, these approaches rely on random or heuristic sampling, which may not balance points across the decision boundary, especially for test instances far from it. To better capture the decision boundary, a local model requires balanced data points on both sides. 

\begin{figure}[htbp]
  \centering
  \scriptsize
  \includegraphics[width=\linewidth, height=0.28\textheight, keepaspectratio]{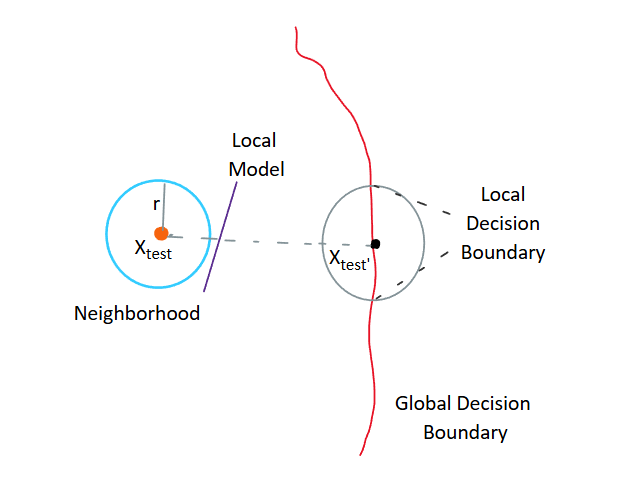}
  \caption{In this example, \(x_{\text{test}}\) is located far away from the global decision boundary. The local model approximates the behavior of the target model in a small neighborhood of  \(x_{\text{test}}\), in terms that it produces the same classifications for the neighborhood data points as the target model. However, the local model (purple line) does not align with the local decision boundary (red curve) of the target model for \(x_{\text{test}}\).}

  \label{fig:local_desc}
\end{figure}

To address these limitations, we propose a novel approach, \textbf{Adversarially Bracketed Local Explanation (ABLE)} , for constructing local models with high fidelity and stability. The key insight of our approach is to explicitly generate data points that bracket the \emph{local decision boundary} for \(x_{\text{test}}\). We emphasize that the local decision boundary for \(x_{\text{test}}\) is a small segment of the target decision boundary that surrounds the projection of \(x_{\text{test}}\) on the decision boundary. In other words, the local decision boundary for \(x_{\text{test}}\) is part of the target decision boundary that is nearest to and thus has the most influence on \(x_{\text{test}}\). The data points generated by our approach lie very close to but on opposite sides of the local decision boundary for \(x_{\text{test}}\). By training a local model on these points, we explicitly target the local decision boundary, resulting in a more stable and faithful representation. Figure~\ref{fig:local_desc} shows that without bracketing, a local model could have excessive flexibility, allowing it to match the target model’s predictions without truly aligning with the target decision boundary.  

Specifically, our ABLE approach consists of three major steps. First, we sample the points in a small neighborhood of \(x_{\text{test}}\) by adding bounded Gaussian noise to \(x_{\text{test}}\). Second, for each neighborhood point \(D\), we construct a pair of adversarial points \(A\) and \(A'\). An adversarial point is a minimally perturbed version of another data point that causes the model to produce a different prediction. We create the pair using a two-step procedure: (i) we apply a (forward) adversarial attack to \(D\) to produce \(A\), which lies on the opposite side of the local decision boundary than \(D\), and (ii) we perform a (reverse) attack on \(A\) to obtain \(A'\), which lies on the opposite side of the local decision boundary than A but the same side as \(D\). Third, we train a local model using these adversarial pairs to approximate the local decision boundary for \(x_{\text{test}}\).  

We evaluate the effectiveness of our approach on six benchmark datasets: Credit Default \cite{credit_default}, Adult Income~\cite{adult_2}, Breast Cancer~\cite{breast_cancer}, Mushroom~\cite{mushroom_73}, Car Evaluation~\cite{car_evaluation_19}, and Covertype~\cite{covertype_31}—using three distinct neural network architectures (MLP, TabNet~\cite{arik2021tabnet}, and TabTransformer~\cite{huang2020tabtransformer}) as target models, resulting in a total of eighteen target models. Moreover, for each of these eighteen models we test all four ABLE variants (ABLE\_PGD, ABLE\_FGSM, ABLE\_etDF, and ABLE\_HSJ)), for a comprehensive evaluation of how different adversarial‐pair generation approaches impact fidelity and stability. Across these 18 configurations, ABLE achieved the highest fidelity in 16 cases, outperforming all baselines including LIME~\cite{ribeiro2016should}, GLIME~\cite{tan2024glime}, USLIME~\cite{saadatfar2024us}, and CALIME~\cite{cinquini2024causality}. In terms of stability, ABLE approaches ranked best in 12 out of 18 cases, with 4 cases resulting in ties with the top-performing baseline.

Our major research contributions are summarized as follows:
\begin{enumerate}[itemsep=0pt, topsep=0pt, leftmargin=*]
    \item \textbf{Approach:} We propose a novel approach to constructing local models for explaining model predictions. Our work is the first to apply adversarial pairs as anchors for local model construction, leading to improved fidelity and stability over random sampling. 
    \item \textbf{Evaluation:} We performed an experimental evaluation of our approach. The results indicate that our approach can construct local models of higher fidelity and stability than the state-of-the-art. 
    \item \textbf{Tool}: We implemented our approach in a publicly available tool, which can be accessed at \cite{anonymous2025able}.
\end{enumerate}

We note that our approach can be considered as a general approach for constructing local models of high fidelity and stability.

\section{Related Work}

The need to explain machine learning models has produced two paths: global approaches that reveal dataset-level patterns and local approaches that detail individual predictions. Global approaches, including feature importance scores \cite{doshi2017towards}, partial dependence plots \cite{christoph2020interpretable}, and model distillation \cite{lipton2018mythos, hinton2015distilling}—capture broad trends but do not explain why a model makes individual decisions. Local explanation approaches, including ours, focus on explaining individual decisions by approximating the model behavior around individual instances \cite{ribeiro2016should, koh2017understanding}. 

LIME~\cite{ribeiro2016should} introduced the idea of training simple local models on perturbed data points near a target instance. However, LIME’s reliance on random perturbations leads to instability, as many data points miss critical regions of the decision boundary \cite{smilkov2017smoothgrad, ghorbani2019interpretation}. GLIME \cite{tan2024glime} improved this approach by weighting perturbations to favor regions near the decision boundary. Still, it relies on stochastic sampling. More importantly, they focus on the test instance, in terms of weighted sampling that favors data points close to the test instance. The decision boundary of the local model they construct does not necessarily align with the target decision boundary. CALIME~\cite{cinquini2024causality} improves fidelity by generating synthetic neighborhoods that respect causal relationships in tabular data. The causal relationships are inferred from a reference dataset, which may not be readily available in practice. USLIME~\cite{saadatfar2024us} leverages the target model’s prediction confidence to drive sampling: points near the model’s uncertainty regions are preferentially chosen, improving boundary coverage. Despite this focus, USLIME still uses heuristic selection criteria and may undersample regions if uncertainty estimates are poor. In contrast, our approach directly focuses on the target decision boundary by generating adversarial pairs that bracket the local decision boundary for the test instance. We then train a local model using those pairs, leading to higher stability and fidelity.

Another line of research improves LIME by utilizing training data to improve the quality of perturbations. For example, AutoLIME~\cite{shankaranarayana2019alime} trains an autoencoder on the training set and performs perturbations in the latent space, thereby generating synthetic data points that better capture the data distribution. Similarly, DLIME~\cite{zafar2019dlime} uses the training data to deterministically select local data points via hierarchical clustering and k-nearest neighbors, which reduces the instability inherent in random perturbations. In contrast, our approach relies solely on model queries to generate adversarial pairs that bracket the decision boundary—a key advantage when training data are unavailable.

SHAP~\cite{lundberg2017unified} contributes to the local explanation field by utilizing Shapley values from game theory to assign feature attributions. While SHAP’s theoretical guarantees are strong, computing exact Shapley values is very expensive since it requires evaluating all possible subsets of features. KernelSHAP approximates Shapley values by sampling feature subsets, but does not explicitly capture the decision boundary of the target model~\cite{lundberg2017unified}.

An alternative perspective is to use inherently interpretable models for explanations~\cite{rudin2019stop}. Rather than relying on post-hoc approaches to explain a complex model's behavior, these approaches advocate designing models that are transparent by nature~\cite{rudin2019stop, ustun2016supersparse}. Models such as rule-based systems ~\cite{wang2015falling}, decision trees~\cite{breiman1984classification}, and generalized additive models~\cite{hastie1990generalized} integrate interpretability into their very structure. However, while these models are inherently interpretable, they struggle to capture more complex relationships in high-dimensional or non-linear data \cite{lipton2018mythos}.

Research on adversarial perturbations~\cite{szegedy2013intriguing, goodfellow2015explaining, madry2018towards} showed that small changes can move a model’s input across the decision boundary. Recent studies~\cite{ghorbani2019interpretation, chen2018learning} use adversarial examples to locate boundary regions, although their main goal is vulnerability detection rather than explanation. Adversarial perturbations have also been used in other areas such as fairness testing \cite{zhang2018mitigating} and robustness training~\cite{shafahi2019adversarial}. In our work, we generate adversarial pairs that straddle the local decision boundary and use them as anchor points to train a local model for explaining individual model decisions.

\section{Background}

In this section, we review the basic concepts in local explanation approaches (with a focus on local models), the key characteristics of effective local explanations, and adversarial data points.

\subsection{LIME-Based Explanations}

LIME \cite{ribeiro2016should} and its variants \cite{tan2024glime,cinquini2024causality,saadatfar2024us} approximate the behavior of a target model with a simpler, interpretable surrogate model in a small neighborhood of a specific instance. At a high level, these approaches perturb \(\mathbf{x}_{\text{test}}\) to generate nearby data points and fits a weighted linear model, where the resulting coefficients indicate feature importance. The quality of the linear model depends to a significant extent on the quality of the sampled data points. Several variants such as GLIME \cite{tan2024glime}, CALIME \cite{cinquini2024causality}, USLIME \cite{saadatfar2024us} and others \cite{zafar2019dlime, shankaranarayana2019alime} have been developed to improve the sampling strategy and thus the quality of the explanations.

A useful local explanation should exhibit two critical properties:

\textbf{Stability:} The explanation should be robust to small perturbations in \(\mathbf{x}_{\text{test}}\) \cite{ribeiro2016should, tan2024glime, doshi2017towards}. This means that adding minor noise or changing the sampling seed should not result in significantly different explanations.

\textbf{Fidelity:} The local model must accurately capture the behavior of the target model in the immediate vicinity of \(\mathbf{x}_{\text{test}}\) \cite{doshi2017towards, tan2024glime}. This ensures that the explanation faithfully represents the model's decision-making process near the instance of interest.

\subsection{Adversarial Data Points}

Given a data point, \( \mathbf{x} \), and a model \(f\), adversarial data points are constructed by identifying the smallest perturbation, \(\delta\), such that:
\[
\mathbf{x}_{\text{adv}} = \mathbf{x} + \delta, \quad \text{with } f(\mathbf{x}_{\text{adv}}) \neq f(\mathbf{x}).
\]
Because \(\delta\) is minimal, these data points lie close to the target decision boundary, revealing the precise location where \(f(\cdot)\) transitions between classes \cite{szegedy2013intriguing, madry2018towards, goodfellow2014explaining}. Adversarial attacks can be untargeted, where the goal is simply to perturb to any class, or targeted, where the goal is to make the model predict a specific different class.

Many adversarial attack methods have been proposed to generate adversarial data points, including Fast Gradient Sign Method (FGSM) \cite{goodfellow2014explaining}, Gradient Descent (PGD) \cite{madry2017towards}, DeepFool~\cite{moosavi2016deepfool}, and ET-DeepFool~\cite{labib2025tailoring}. Our approach is agnostic to the attack method and can work with any of them, as demonstrated in Section \ref{sec:exp-design}.

\section{Motivation}

Although LIME is popular for its simplicity and model-agnostic nature, it suffers from two major shortcomings, as highlighted in \cite{tan2024glime}.

\paragraph{Instability}  
LIME generates local perturbations by randomly sampling points around the test instance \( x_{\text{test}} \), resulting in varying sample sets and inconsistent explanations. Additionally, LIME assigns weights to each perturbation based on an exponential kernel function defined as 
$
\exp \left( -\frac{\| x - x_{\text{test}} \|_2^2}{\sigma^2} \right),
$
where \( \sigma \) is the kernel width controlling the locality of explanations. Tan et al. \cite{tan2024glime} showed that small \( \sigma \) values (e.g., 0.25), sharply reduce weights for slightly distant points, diminishing their influence and increasing instability.

\begin{figure}[htbp]
  \centering
  \includegraphics[width=\linewidth, height=0.3\textheight, keepaspectratio]{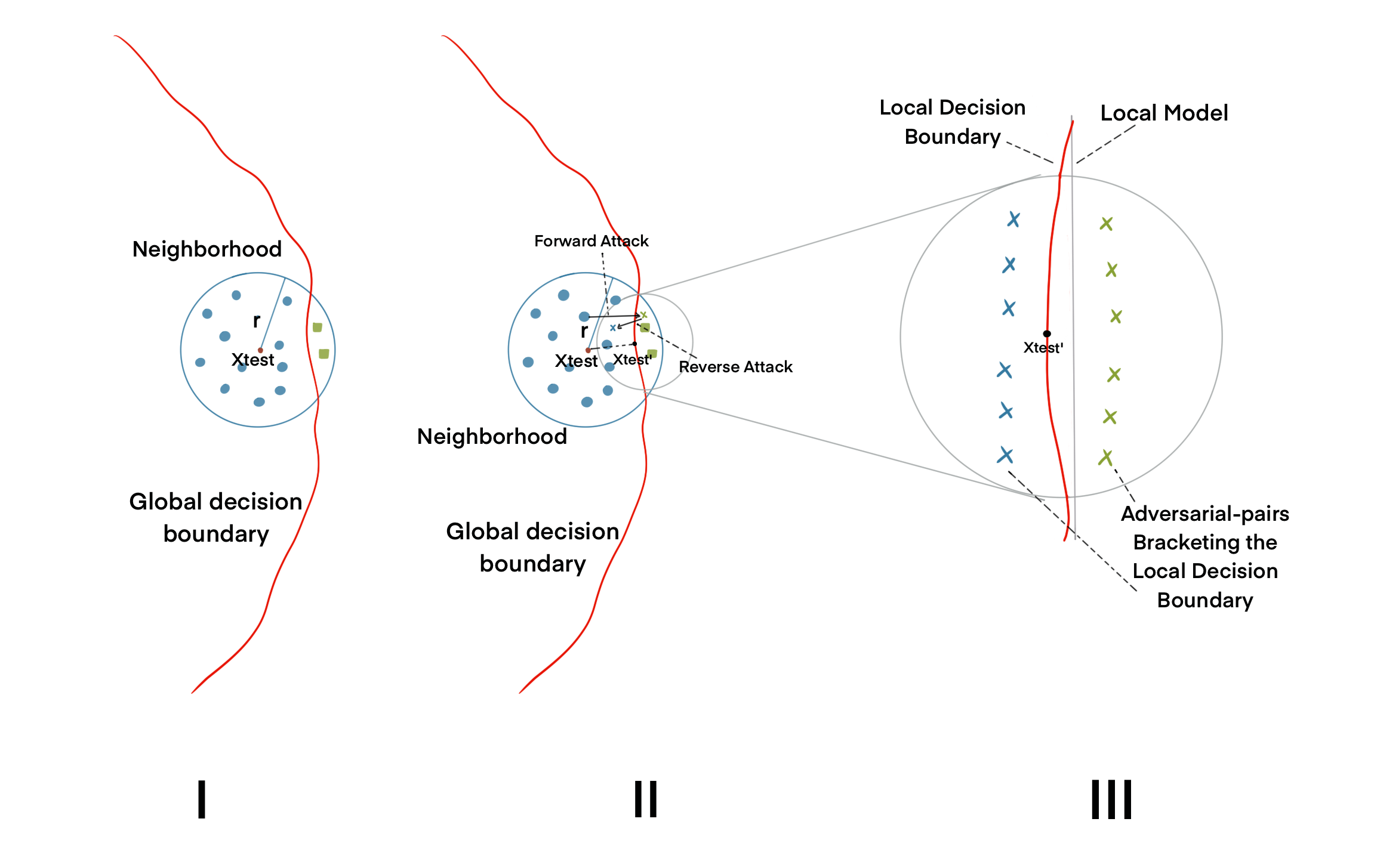}
  \caption{(I) \(X_{test}\) is surrounded by neighborhood points (blue dots for same class and green boxes for different class) within radius \(r\) (red: global decision boundary). (II) \(X_{test'}\) is the projection onto the decision boundary; forward and reverse adversarial attacks generate adversarial pairs. (III) Adversarial pairs (blue crosses for the same class and green crosses for a different class) bracket the local decision boundary and are used to train the local model.}
  \label{fig:three_images}
\end{figure}

\paragraph{Poor Fidelity}  
For a local model to be faithful, it must capture the target model’s behavior where predictions change— near the decision boundary \cite{chen2018learning, ghorbani2019interpretation}. Data points near the decision boundary provide crucial insights into how small changes in input lead to label flips. LIME’s broad, unguided sampling tends to perturb \(\mathbf{x}_{\text{test}}\) randomly in all directions. Furthermore, LIME primarily samples data points from the neighborhood around \(x_{\text{test}}\) and its proximity weighting scheme favors those data points. If \(x_{\text{test}}\) is far from the decision boundary, the sampled points in its vicinity are also likely to be far from the decision boundary (see Figure~\ref{fig:local_desc}).  This can cause the local model to misrepresent the local decision-making process.

The major challenge in building a local model of high fidelity and stability is how to faithfully replicate the decision boundary of the target classifier. As shown in Figure~\ref{fig:three_images}.I, when a test instance \(X_{test}\) (shown in orange) is perturbed within a neighborhood of radius \(r\), the majority of the data points (blue dots) may fall on the same side of the decision boundary (red line), while only a few (green dots) lie on the opposite side. If \(X_{test}\) is far from the decision boundary, all neighborhood data points may belong to one side (e.g., see Figure~\ref{fig:local_desc}). Such imbalanced sampling leads to incomplete coverage of the decision boundary. Without enough data points from both sides of the boundary, the local model is trained on an incomplete view of the decision boundary region. As a result, it may not faithfully replicate the decision boundary.

To overcome this challenge, we explicitly target the space around the \emph{local decision boundary} that most influences \(X_{test}\). We generate pairs of data points that bracket this boundary. As illustrated in Figure~\ref{fig:three_images}.II, for each neighborhood instance, a forward PGD attack minimally perturbs the point until its predicted label flips, effectively pushing it to the other side of the decision boundary. A reverse PGD attack then perturbs the flipped point back to its original side. This two-step process creates an adversarial pair that tightly brackets the target boundary.

Figure~\ref{fig:three_images}.III  further shows the result of our approach. Here, adversarial pairs (marked with \(X\) symbols) are shown for multiple neighborhood data points. This targeted sampling strategy ensures that the local model is trained on data that covers both sides of the critical decision boundary region, resulting in a robust local approximation of the classifier’s behavior.

\section{Approach}
\label{sec:approach}

In this section, we present our \textbf{Adversarially Bracketed Local Explanation (ABLE)} approach for constructing a local model around a specific test instance $\mathbf{x}_{\text{test}}$. The fundamental idea is to generate a set of adversarial pairs that bracket the target decision boundary rather than relying on random sampling. We first describe the approach for binary classification, then discuss the multi-class extension in a separate subsection.

\subsection{Problem Setup and Notation}
\label{subsec:problem-setup}

Let $f: \mathbb{R}^d \to \{0, 1\}$ be a binary classifier that assigns one of two classes to any input in $\mathbb{R}^d$, where $d$ is the feature dimensionality. We aim to explain $f(\mathbf{x}_{\text{test}})$ by training a local model $g(\mathbf{x})$ in the neighborhood of radius $r$ around $\mathbf{x}_{\text{test}}$:
\[
\mathcal{N}(\mathbf{x}_{\text{test}}, r)
= \left\{\mathbf{x} \in \mathbb{R}^d \,\middle|\, \|\mathbf{x} - \mathbf{x}_{\text{test}}\|_2 \le r \right\}.
\]
We assume that the target model $f(\cdot)$ allows us to retrieve prediction probabilities for any input $\mathbf{x}$. Our local model is an interpretable logistic regression classifier of the form:
\[
g(\mathbf{x}) = \sigma(\mathbf{w}^\top \mathbf{x} + b),
\]
where $\sigma(z) = \frac{1}{1 + e^{-z}}$ is the sigmoid function, and $(\mathbf{w}, b)$ are the model parameters. We train $g(\cdot)$ to approximate the decision boundary near $\mathbf{x}_{\text{test}}$, and interpret the learned coefficients to understand local feature importance. The multi-class extension is discussed in Subsection~\ref{subsec:multi-class}.

\subsection{Overview of Our Approach}

Our approach, ABLE, has three main steps:
\begin{enumerate}[itemsep=0pt, topsep=0pt, leftmargin=*]

\item \textbf{(Step 1) Neighborhood Generation:} We sample a set of points $\{\mathbf{x}_i\}_{i=1}^n$ in the vicinity of the test point $\mathbf{x}_{\text{test}}$ by adding bounded random noise to $\mathbf{x}_{\text{test}}$.
    
\item \textbf{(Step 2) Adversarial Pair Generation:} For each $\mathbf{x}_i$, we apply an adversarial attack method to find a minimal perturbation that changes the classifier's prediction, resulting in an adversarial point $\mathbf{x}_i^{\text{adv}}$. We then apply a reverse perturbation process to return $\mathbf{x}_i^{\text{adv}}$ to its original label, forming a pair $(\mathbf{x}_i^{\text{adv}}, \mathbf{x}_i^{\text{rev}})$. These two points lie on opposite sides of the decision boundary.

\item \textbf{(Step 3) Local Model Training:} We train a logistic regression model on these adversarial pairs to capture the local decision boundary.
\end{enumerate}

Each neighborhood point $\mathbf{x}_i$ generates an adversarial pair that brackets the local decision boundary, providing boundary-focused coverage. In the following subsections, we detail each step.

\subsection{Step 1: Neighborhood Generation}

We first generate a set of $n$ data points in the neighborhood of the test point $\mathbf{x}_{\text{test}}$. For each sample, we draw a noise vector $\mathbf{z}_i$ from a standard multivariate Gaussian distribution, $\mathbf{z}_i \sim \mathcal{N}(\mathbf{0}, \mathbf{I})$. To ensure that the perturbation remains within a prescribed radius $r$, we rescale the noise by computing
\[
\delta_i = \mathbf{z}_i \cdot \min\left(1,\,\frac{r}{\|\mathbf{z}_i\|_2}\right),
\]
ensuring that $\|\delta_i\|_2 \le r$. The neighborhood sample is then obtained by adding this perturbation to the original test point, i.e., $\mathbf{x}_i = \mathbf{x}_{\text{test}} + \delta_i$, and the corresponding label is determined by evaluating the classifier $f$ at $\mathbf{x}_i$, so that $y_i = f(\mathbf{x}_i)$. This process results in the neighborhood dataset
$
\mathcal{D}_{\text{neighbors}} = \left\{ (\mathbf{x}_i,\, y_i) \right\}_{i=1}^n.
$
We include the original test point $\mathbf{x}_{\text{test}}$ in this neighborhood dataset. Unlike pure random-sampling approaches, our method uses each $\mathbf{x}_i$ as a starting point for generating adversarial pairs. These pairs lie on opposite sides of the target decision boundary, effectively bracketing the boundary and capturing the critical transition region.

\subsection{Step 2: Adversarial Pair Generation}

For each neighborhood point $(\mathbf{x}_i, y_i)$ from Step~1, we generate an adversarial pair that tightly brackets the classifier’s decision boundary. First, starting from $\mathbf{x}_i$, we apply an adversarial attack to obtain
\[
\mathbf{x}_i^{\mathrm{adv}} = \mathbf{x}_i + \delta_i^{\mathrm{adv}},
\]
such that $f(\mathbf{x}_i^{\mathrm{adv}}) \neq y_i$. To find the minimal perturbation $\delta_i^{\mathrm{adv}}$, we begin with a conservative default budget. We then execute the attack, whether single-step or iterative. If no adversarial example is found, the budget is increased incrementally (e.g., by 0.1) until success. Attack methods that automatically compute minimal perturbations determine $\delta_i^{\mathrm{adv}}$ without a preset bound.

Next, from $\mathbf{x}_i^{\mathrm{adv}}$, we apply a second adversarial attack to return to the original class:
\[
\mathbf{x}_i^{\mathrm{rev}} = \mathbf{x}_i^{\mathrm{adv}} + \delta_i^{\mathrm{rev}},
\]
such that $f(\mathbf{x}_i^{\mathrm{rev}}) = y_i$. When there are only two classes, flipping the prediction from $\mathbf{x}_i^{\mathrm{adv}}$ naturally returns to $y_i$. This ensures that the resulting pair brackets the decision boundary. Each neighborhood point thus results in one adversarial pair $(\mathbf{x}_i^{\mathrm{adv}}, \mathbf{x}_i^{\mathrm{rev}})$.

\subsection{Step 3: Local Model Training}

After generating adversarial pairs for all $n$ neighborhood points, we obtain
\[
\mathcal{D}_{\mathrm{adv}} = \left\{(\mathbf{x}_i^{\mathrm{adv}}, f(\mathbf{x}_i^{\mathrm{adv}})),\, (\mathbf{x}_i^{\mathrm{rev}}, y_i)\right\}_{i=1}^n,
\]
where $\mathbf{x}_i^{\mathrm{adv}}$ is labeled with $f(\mathbf{x}_i^{\mathrm{adv}})$, and $\mathbf{x}_i^{\mathrm{rev}}$ with the original $y_i = f(\mathbf{x}_i)$.

We then train a binary logistic regression model by minimizing the cross-entropy loss on $\mathcal{D}_{\mathrm{adv}}$. Because the adversarial examples lie close to the true decision boundary, this linear surrogate closely approximates the local behavior of $f$. 

\subsection{Multi-Class Extension}
\label{subsec:multi-class}

For multi-class classification, where $f: \mathbb{R}^d \to \{1, 2, \dots, C\}$ with $C > 2$, the approach requires minor adaptations. Steps 1 and 3 are largely unchanged, but Step 2 and the local model formulation are adjusted as follows.

In Step 2, the initial adversarial attack remains untargeted, yielding $\mathbf{x}_i^{\mathrm{adv}}$ with $f(\mathbf{x}_i^{\mathrm{adv}}) \neq y_i$. However, for the reverse attack, we use a targeted attack to explicitly direct the perturbation back to the original class $y_i$, as an untargeted flip might land in a third class. This ensures the pair brackets the boundary between $y_i$ and the perturbed class.

In Step 3, we replace binary logistic regression with a multinomial logistic (softmax) model~\cite[Eq.~(8.33)]{Murphy2012}:
\[
g_c(\mathbf{x}) =
\frac{\exp\left(\mathbf{w}_c^\top \mathbf{x} + b_c\right)}
     {\sum_{k=1}^C \exp\left(\mathbf{w}_k^\top \mathbf{x} + b_k\right)},
\quad c = 1,\dots,C,
\]
and fit the parameters $(W, b)$ by minimizing the cross-entropy on $\mathcal{D}_{\mathrm{adv}}$.

\begin{table}[htbp]
\centering
\caption{The table summarizes the number of instances, categorical and continuous features, and class distributions for the datasets.}
\small
\label{tab:datasets}
\resizebox{\columnwidth}{!}{
\begin{tabular}{ccccc}
\toprule
Dataset & Instances & Categorical & Continuous & Classes \\
\midrule
Credit Default & 30,000 & 23 & 0 & 2 \\
Adult Income & 48,842 & 8 & 6 & 2 \\
Breast Cancer & 569 & 0 & 30 & 2 \\
Mushroom & 8,124 & 22 & 0 & 2 \\
Car Evaluation & 1,728 & 6 & 0 & 4 \\
Covertype & 581,012 & 10 & 44 & 7 \\
\bottomrule
\end{tabular}
}
\end{table}

\section{Experiment Design}

\label{sec:exp-design}

In this section, we present the experimental setup used to evaluate our proposed approach, \emph{ABLE}, in constructing local models. Our primary objectives are:
\begin{enumerate}[itemsep=0pt, topsep=0pt, leftmargin=*]
    \item \textbf{RQ1: Fidelity.} How accurately does ABLE replicate the target model’s behavior, as measured by the \(R^2\) score between the target and local model outputs?
    \item \textbf{RQ2: Stability.} How consistent are ABLE’s explanations, i.e., important features, under small perturbations of the test instance, as measured by the Jaccard Index on the top-\(K\) features?
    \item \textbf{RQ3: Hyperparameter Impact.} How do variations in the neighborhood radius and number of adversarial pairs affect ABLE’s performance?
\end{enumerate}

\subsection{Datasets}
We evaluate ABLE on six benchmark datasets from the UCI Machine Learning Repository, commonly used in tabular classification tasks: Credit Default \cite{credit_default}, Adult Income \cite{adult_2}, Breast Cancer \cite{breast_cancer}, Mushroom \cite{mushroom_73}, Car Evaluation~\cite{car_evaluation_19}, and Covertype \cite{covertype_31}. These datasets vary in size, feature types, and class distributions, as summarized in Table~\ref{tab:datasets}. To preprocess the data, we use the RDT HyperTransformer \cite{rdt} to transform both categorical and continuous features into a unified numerical representation while preserving the original number of features. The HyperTransformer automatically detects feature types and applies appropriate transformers: categorical features are numerically encoded using methods such as label encoding, which maps each category to a unique integer, and continuous features are scaled using transformers like FloatFormatter or GaussianNormalizer. Missing values are handled using strategies such as mean imputation or random sampling, as configured by the HyperTransformer. The transformed features are then standardized using a StandardScaler to achieve zero mean and unit variance, ensuring compatibility with downstream model training and evaluation. Each dataset is split into training (70\%), validation (15\%), and test (15\%) sets, following standard practices in tabular machine learning \cite{kuhn2013applied}.

\begin{table}[htbp]
\centering
\caption{Test accuracy of target models on 15\% test data, trained on 70\% train data with 15\% validation split.}
\label{tab:model_performance}
\resizebox{\columnwidth}{!}{
\begin{tabular}{lccc}
\toprule
\multirow{2}{*}{\textbf{Dataset}} & \multicolumn{3}{c}{\textbf{Model Accuracy}} \\
\cmidrule(lr){2-4}
 & \textbf{MLP } & \textbf{TabNet } & \textbf{TabTransformer } \\
\midrule
Adult Income & 0.845 & 0.834 & 0.847 \\
Breast Cancer & 0.982 & 0.930 & 0.930 \\
Credit & 0.816 & 0.817 & 0.817 \\
Mushroom & 0.991 & 0.994 & 0.988 \\
Car Evaluation & 0.846 &  0.876  & 0.715 \\
Covertype & 0.815 & 0.884  &  0.921  \\
\bottomrule
\end{tabular}
}
\end{table}

\subsection{Target Models}
We train three neural network classifiers for each dataset to serve as target models:
\begin{enumerate}[itemsep=0pt, topsep=0pt, leftmargin=*]
     
   \item \textbf{MLP Classifier.} A fully connected neural network with two hidden layers, each followed by ReLU activation.

   \item \textbf{TabNet Classifier~\cite{arik2021tabnet}.} A sequential attention-based model designed for tabular data, leveraging feature-wise dynamic attention to identify relevant patterns.

   \item \textbf{TabTransformer Classifier~\cite{huang2020tabtransformer}.} A transformer-based model with multi-head self-attention for feature encoding in tabular data.

\end{enumerate}
Each model is trained on the training set, tuned on the validation set, and evaluated on the test set. Test accuracy for each model is reported in Table~\ref{tab:model_performance}.

\subsection{Adversarial Sample Generation}

We generate adversarial pairs using four attack variants: PGD~\cite{madry2017towards}, FGSM~\cite{goodfellow2014explaining}, an enhanced DeepFool variant (etDF)~\cite{labib2025tailoring}, and HopSkipJump (HSJ)~\cite{chen2020hopskipjumpattack}. PGD, FGSM, and etDF are white-box, gradient-based attacks, while HSJ is a black-box, query-based method that does not require access to model internals. All attacks are implemented using the Adversarial Robustness Toolbox (ART)~\cite{art}, except etDF, which uses the authors’ official implementation~\cite{labib2025tailoring}.

For each test instance, we sample a neighbor \(\mathbf{x}_i\) and apply the chosen attack to generate an adversarial point \(\mathbf{x}_i^{\mathrm{adv}}\) such that \(f(\mathbf{x}_i^{\mathrm{adv}}) \neq y_i\). The procedure varies by each attack method:

\begin{enumerate}[itemsep=0pt, topsep=0pt, leftmargin=*]

    \item \textbf{ABLE\_FGSM:} Applies a single-step gradient update using ART’s default \(\epsilon\), which defines the radius of the perturbation ball. If no adversarial example is found, \(\epsilon\) is incremented by 0.1 until success.

    \item \textbf{ABLE\_PGD:} A multi-step variant that builds on FGSM by performing iterative gradient-based updates within the same \(\epsilon\)-ball. If no adversarial example is generated, \(\epsilon\) is similarly increased by 0.1.

    \item \textbf{ABLE\_etDF:} Uses an adaptive DeepFool procedure to compute the minimal perturbation needed to reach the decision boundary, without requiring a fixed \(\epsilon\).

    \item \textbf{ABLE\_HSJ:} A black-box attack that estimates the decision boundary via query-based sampling and refines the perturbation using binary search.
\end{enumerate}

\subsection{Hyperparameters}
To investigate the impact of hyperparameters on the effectiveness of our approach, we conducted an analysis on the following two key hyperparmeters configurations:

\begin{enumerate}[itemsep=0pt, topsep=0pt, leftmargin=*]
    \item \textbf{Neighborhood Radius (r):} This parameter defines the local region around \(x_{\text{test}}\) from which the neighborhood points are sampled.  We vary $r$ across the values \{0.2, 0.4, 0.6, 0.8, 1.0\}. 
    \item \textbf{Number of Neighborhood Points (\(n\)):} We vary \(n\), the number of points sampled within \(\mathcal{N}(x_{\text{test}}, r)\). We vary $n$ across the values \{50, 75, 100, 150\}
\end{enumerate}

For each configuration, we generate adversarial pairs, train the local model, and evaluate its fidelity.

\subsection{Baselines}
We compare \emph{ABLE} against four local explanation baselines:
\begin{enumerate}[itemsep=0pt, topsep=0pt, leftmargin=*]

    \item \textbf{LIME}~\cite{ribeiro2016model}. Uses random perturbations and a weighted linear regressor to approximate the local decision boundary of a target model.  
    \item \textbf{GLIME}~\cite{tan2024glime}. Improves LIME stability by integrating the weighting function into the sampling distribution, reducing the reliance on post-hoc weighting. 
    \item \textbf{USLIME}~\cite{saadatfar2024us}.  Extends LIME by using uncertainty sampling to select neighborhood points close to the decision boundary, based on a boundary score from prediction probabilities and distance filtering, improving fidelity on tabular data.
    
    \item \textbf{CALIME}~\cite{cinquini2024causality}. Enhances LIME by generating causal-aware neighborhoods using NCDA and GENCDA, improving fidelity and stability. It requires a reference dataset for causal discovery. In this experiment, we use the training dataset as the reference dataset. We note that a reference dataset may not be available in practice. Other methods do not require a reference dataset.

\end{enumerate}

\begin{table}[htbp]
\centering
\small
\caption{Effect of Neighborhood Radius (\(r\)) and Number of Neighborhood Points (\(n\)) on Local Fidelity. Higher values indicate better fidelity.}

\label{tab:surrogate_fidelity}
\resizebox{\columnwidth}{!}{
\begin{tabular}{ccccc}
\toprule
\multirow{2}{*}{\shortstack{\textbf{Neighborhood} \\ \textbf{Radius} (\(r\))}} & \multicolumn{4}{c}{\textbf{Number of Neighborhood Points} (\(n\))} \\
\cmidrule(lr){2-5}
 & \textbf{50} & \textbf{75} & \textbf{100} & \textbf{150} \\
\midrule
0.2 & 0.840 & 0.870 & 0.889 & 0.908 \\
0.4 & 0.834 & 0.862 & 0.881 & 0.904 \\
0.6 & 0.834 & 0.863 & 0.880 & 0.902 \\
0.8 & 0.829 & 0.857 & 0.875 & 0.896 \\
1.0 & 0.827 & 0.854 & 0.870 & 0.890 \\
\bottomrule
\end{tabular}
}
\label{ntable}
\end{table}

\begin{table*}[htbp]
\centering
\small
\caption{Fidelity Comparison. Higher \textbf{R\textsuperscript{2}} indicates better alignment with the target model.}

\begin{tabular}{l l c c c c c c c c}
\toprule
Dataset & Classifier & ABLE\_PGD & ABLE\_etDF & ABLE\_FGSM & ABLE\_HSJ & LIME & GLIME & USLIME & CALIME \\
\midrule
\multirow{3}{*}{Credit} 
& MLP & \textbf{0.983} & 0.964 & 0.900 &  0.948  & 0.412 & 0.407 & 0.361 & 0.313 \\
& TabNet & 0.759 & \textbf{ 0.866} & 0.837 & 0.723   & 0.831 & 0.845 & 0.730 & 0.661 \\
& TabTransformer & 0.954 & \textbf{0.984} & 0.967 & 0.933  & 0.385 & 0.377 & 0.467 & 0.164 \\
\midrule
\multirow{3}{*}{Breast\_Cancer} 
& MLP & \textbf{0.959} & 0.954 & 0.932 & 0.917 & 0.236 & 0.236 & 0.145 & 0.911 \\
& TabNet & \textbf{0.890} & 0.876 & 0.816 & 0.731  & 0.388 & 0.380 & 0.337 & 0.888 \\
& TabTransformer & \textbf{0.979} & 0.977 & 0.939  & 0.895 & 0.501 & 0.501 & 0.356 & 0.879 \\
\midrule
\multirow{3}{*}{Mushroom} 
& MLP & 0.975 & \textbf{0.991} & 0.971 & 0.942  & 0.373 & 0.350 & 0.465 & 0.893 \\
& TabNet & 0.888 & \textbf{0.932} & 0.892 &  0.697  & 0.588 & 0.594 & 0.495 & 0.872 \\
& TabTransformer & \textbf{0.975} & 0.959 & 0.965 & 0.922  & 0.394 & 0.393 & 0.215 & 0.753 \\
\midrule
\multirow{3}{*}{Covertype} 
& MLP & \textbf{0.972} & 0.911 & 0.961 &  0.863   &  0.026    & 0.023 & 0.133 & 0.559 \\
& TabNet & \textbf{0.863} & 0.749 & 0.788 &  0.702  &  0.505 & 0.510 & 0.422 & 0.531 \\
& TabTransformer & \textbf{0.898} & 0.881 & 0.868 & 0.701  & 0.233 & 0.229 & 0.176 & 0.460 \\
\midrule
\multirow{3}{*}{Adult} 
& MLP & 0.932 & \textbf{0.952} & 0.928 & 0.842  & 0.738 & 0.757 & 0.858 & 0.562 \\
& TabNet & 0.752 & 0.874 & 0.722 &  0.744 & \textbf{0.946} & 0.945 & 0.884 & 0.456 \\
& TabTransformer & \textbf{0.907} & 0.872 & 0.892 & 0.733 & 0.554 & 0.572 & 0.670 & 0.358 \\
\midrule
\multirow{3}{*}{Car} 
& MLP & \textbf{0.986} & 0.966 & 0.961 & 0.959  &  0.721 & 0.665 & 0.896 & 0.719 \\
& TabNet & 0.817 & 0.728 & 0.845 & 0.853  & 0.708 & 0.725 & \textbf{0.907} & 0.551 \\
& TabTransformer & \textbf{0.826} & 0.728 & 0.657 & 0.926   & 0.489 & 0.530 & 0.779 & 0.489 \\
\midrule
\multicolumn{2}{l}{Average} & \textbf{0.906} & 0.898 & 0.880 &  0.831  & 0.502 & 0.502 & 0.516 & 0.612 \\
\bottomrule
\end{tabular}
\label{tab:fidelity}
\end{table*}

\subsection{Evaluation Metrics}

\paragraph{1) Fidelity ($R^2$ Score).}
We measure how well the predicted probabilities of the local model match the probabilities of the target model in a local neighborhood of \emph{test} instances. For each $x_{\text{test}}$, we generate a set of neighborhood points by adding bounded Gaussian noise, then split these points into training and evaluation sets. The local model is trained on adversarial pairs constructed from the training set, and its fidelity is evaluated on the separate evaluation set. For baseline approaches, the evaluation set is drawn similarly from their respective local neighborhoods using the prescribed sampling procedure for each method.  We compute the coefficient of determination ($R^2$) between the output of the local model and the target model in this evaluation set:
\[
R^2(f, g) 
= 
1 
\;-\; 
\frac{
  \sum_{x \in x_{\text{test}}}(f(x) - g(x))^2
}{
  \sum_{x \in x_{\text{test}}}(f(x) - \bar{f})^2
},
\]
where \( f(x) \) is the target model’s probability estimate, and \( g(x) \) is the local model’s estimate. A higher \( R^2 \) indicates better fidelity \cite{bishop2006pattern}, which means that the local model more accurately approximates the predictions of the target model \cite{bishop2006pattern} in the neighborhood.

\paragraph{2) Stability (Jaccard Index).}
We assess the stability of each explanation by introducing a small random noise to the test instance to create a single perturbed version. For both the original and perturbed instances, we generate explanations and extract their top-$K$ important features (with \(K=5\) being commonly used in the literature). We then compute the Jaccard Index between these two sets of top-$K$ features:
\[
\text{JI}(\mathcal{F}_{\text{original}}, \mathcal{F}_{\text{perturbed}})
=
\frac{|\mathcal{F}_{\text{original}} \cap \mathcal{F}_{\text{perturbed}}|}{|\mathcal{F}_{\text{original}} \cup \mathcal{F}_{\text{perturbed}}|},
\]
where $\mathcal{F}_{\text{original}}$ and $\mathcal{F}_{\text{perturbed}}$ denote the sets of top-$K$ features from the original and perturbed instances, respectively. A higher Jaccard Index indicates that the explainer consistently identifies the same features as important across the original and slightly perturbed instances \cite{tan2005introduction}.

\paragraph{3) Time.}
We report the total run time in seconds for each approach to generate a local model, from sample/perturbation generation to final model training. Lower time indicates higher efficiency.

\subsection{Experimental Procedure}
We evaluate our approach by first training three target models (MLP, TabNet, and TabTransformer) on the training set, validating on the validation set, and fixing the final models for testing. We select 100 random test instances per dataset and evaluate ABLE with each adversarial attack method alongside the baseline methods. We then assess the fidelity and stability. In addition, we record the total computation time for each method. This entire process is repeated over 10 random seeds, and the resulting \(R^2\), Jaccard Index, and runtime metrics are averaged.

All experiments were performed on a machine with a multi-core processor operating at 2.10 GHz and 32 GB of RAM.

\begin{table*}[htbp]
\centering
\small
\caption{Stability Comparison. Higher Jaccard Index indicates more consistent feature importance under perturbations.}

\begin{tabular}{l l c c c c c c c c}
\toprule
Dataset & Classifier & ABLE\_PGD & ABLE\_etDF & ABLE\_FGSM  &  ABLE\_HSJ  & LIME & GLIME & USLIME & CALIME \\
\midrule
\multirow{3}{*}{Credit} 
& MLP & \textbf{0.984} & 0.833 & 0.974 & 0.883   & 0.800 & 0.787 & 0.795 & 0.452 \\
& TabNet & \textbf{1.000} & 0.930 & 0.882 &  1.000  & 0.899 & \textbf{1.000} & \textbf{1.000} & 0.429 \\
& TabTransformer & 0.956 & 0.913 & 0.981  &  0.983  & \textbf{1.000} & \textbf{1.000} & 0.982 & 0.313 \\
\midrule
\multirow{3}{*}{Breast\_Cancer} 
& MLP & 0.983 & \textbf{1.000} & 0.984  &  0.867   & 0.902 & 0.917 & 0.562 & 0.357 \\
& TabNet & 0.833 & 0.889 & 0.867 &  0.818  & 0.938 & \textbf{0.963} & 0.834 & 0.215 \\
& TabTransformer & \textbf{0.989} & 0.977 & 0.957  & 0.967   & 0.943 & 0.963 & 0.943 & 0.284 \\
\midrule
\multirow{3}{*}{Mushroom} 
& MLP & 0.913 & \textbf{0.970} & 0.939 & 0.917   & 0.905 & \textbf{0.970} & 0.905 & 0.424 \\
& TabNet & 0.854 & \textbf{0.950} & 0.889 &   0.767  & 0.802 & 0.921 & 0.905 & 0.424 \\
& TabTransformer & 0.926 & 0.931 & 0.931 & 0.850    & \textbf{1.000} & 0.978 & 0.857 & 0.343 \\
\midrule
\multirow{3}{*}{Covertype} 
& MLP & 0.778 & 0.833 & \textbf{1.000} &  0.952   & 0.311 & 0.273 & 0.515 & 0.299 \\
& TabNet & 0.833 & 0.810 & 0.833  &  0.848  & 0.882 & 0.961 & \textbf{0.980} & 0.376 \\
& TabTransformer & 0.714 & 0.833 & \textbf{1.000} & 0.879  & 0.730 & 0.792 & 0.730 & 0.349 \\
\midrule
\multirow{3}{*}{Adult} 
& MLP & 0.956 & 0.963 & 0.960  &  0.967  & 0.955 & \textbf{0.988} & 0.967 & 0.478 \\
& TabNet & 0.952 & 0.965 & 0.939  &  0.848   & \textbf{1.000} & \textbf{1.000} & \textbf{1.000} & 0.621 \\
& TabTransformer & 0.943 & 0.944 & \textbf{0.974}  &  0.846   & 0.882 & 0.793 & 0.937 & 0.498 \\
\midrule
\multirow{3}{*}{Car} 
& MLP & 0.958 & \textbf{0.985} & 0.972 &  0.967  & 0.945 & 0.947 & 0.960 & 0.767 \\
& TabNet & 0.944 & \textbf{1.000} & 0.889   &  0.981   & 0.910 & \textbf{1.000} & \textbf{1.000} & 0.899 \\
& TabTransformer & 0.958 & 0.987 & \textbf{1.000}   &  0.883    & \textbf{1.000} & \textbf{1.000} & 0.972 & 0.796 \\
\midrule
\multicolumn{2}{l}{Average} & 0.915 & 0.928 & \textbf{0.932} & 0.901   & 0.878 & 0.903 & 0.880 & 0.435 \\
\bottomrule
\end{tabular}
\label{tab:stability}
\end{table*}

\begin{table*}[h]
\centering
\caption{Runtime Comparison in Seconds (Aggregated Across All Datasets). Lower values indicate better computational efficiency.}
\small
\begin{tabular}{lrrrrrrrr}
\toprule
Classifier      & ABLE\_etDF & ABLE\_PGD & ABLE\_FGSM & ABLE\_HSJ & LIME  & GLIME & USLIME & CALIME    \\
\midrule
MLP             & 0.416          & 1.367     & 0.121  & 2.124    & 0.111 & 0.153 & 0.163  & 52.345   \\
TabNet          & 1.432          & 3.781    & 0.480  &  4.140  & 0.210 & 0.249 & 0.420  & 52.509   \\ 
TabTransformer  & 0.602          & 2.831     & 0.163  & 2.441   & 0.120 & 0.157 & 0.147  & 64.785   \\
\bottomrule
\end{tabular}
\label{tab:time}
\end{table*}

\section{Results and Discussion}
\label{sec:experiments}

\subsection{Hyperparameter Analysis}

We evaluated the sensitivity of the ABLE framework to its primary hyperparameters: the neighborhood radius (\(r\)) and the number of neighborhood points (\(n\)). Table~\ref{ntable} presents the average local fidelity ($R^2$) across all four ABLE approaches, six datasets, and three classifiers.

\textbf{Neighborhood Radius ($r$)}: Fidelity is highest at a smaller radius (e.g., $r=0.2$), achieving an $R^2$ of up to 0.908. This indicates that a tighter, more localized neighborhood allows the local model to better approximate the target model's behavior. As \(r\) increases, fidelity declines, suggesting that a larger radius introduces points from regions where the linear approximation is less accurate. 

\textbf{Number of Neighborhood Points (\(n\))}: Fidelity consistently improves as \(n\) increases. However, the gains diminish for larger \(n\), while the computational cost rises. Notably, ABLE achieves high fidelity (0.908) with just 150 points ($n=150$).

Based on these findings, we selected \(r = 0.2\) and \(n = 150\) for all subsequent experiments, balancing high performance with computational efficiency.

\subsection{Fidelity Comparison}
Table~\ref{tab:fidelity} presents the fidelity results ($R^2$ scores) comparing ABLE's four variants against the baselines. ABLE consistently and significantly outperforms all baselines in terms of fidelity. On average, ABLE\_PGD achieves the highest fidelity with an $R^2$ score of 0.906, followed closely by ABLE\_etDF (0.898), ABLE\_FGSM (0.880), and ABLE\_HSJ (0.831). In contrast, the baselines perform poorly, with average scores of 0.612 for CALIME and around 0.5 for LIME, GLIME, and USLIME.

This superiority is evident across nearly all configurations. For example, on the Covertype dataset with an MLP target, ABLE\_PGD achieves a near-perfect fidelity of 0.972, whereas LIME and GLIME score a mere 0.026 and 0.023, respectively. This demonstrates that by generating data points that bracket the decision boundary, ABLE creates a local model that more accurately reflects the target model's decision-making process. The only exceptions are two instances (TabNet on Adult and Car) where baselines achieve higher scores, though ABLE's overall performance remains strong.

\subsection{Stability Comparison}

Table~\ref{tab:stability} shows the stability results measured by the Jaccard Index. All ABLE variants demonstrate superior stability on average compared to the baselines. ABLE\_FGSM leads with an average Jaccard Index of 0.932, followed by ABLE\_etDF (0.928), ABLE\_PGD (0.915), and ABLE\_HSJ (0.901). These scores are higher than those of GLIME (0.903), LIME (0.878), USLIME (0.880), and especially CALIME (0.435).

For instance, on the Covertype dataset with an MLP model, ABLE\_FGSM achieves perfect stability (1.000), while LIME and GLIME are highly unstable (0.311 and 0.273). This indicates that by anchoring the local model to the decision boundary, ABLE produces explanations that are more robust to minor input perturbations.

\subsection{Runtime Analysis}

Table~\ref{tab:time} compares the computational time required to generate a single explanation. As expected, ABLE's iterative adversarial search incurs a higher computational cost than fast sampling methods like LIME, GLIME, and USLIME.  ABLE\_FGSM is the fastest variant—comparable to LIME, GLIME, and USLIME—due to its single-step attack, while the iterative nature of PGD makes ABLE\_PGD the slowest. ABLE\_HSJ, being a black-box method, lies in between but is slower than PGD due to repeated boundary estimations. However, ABLE remains significantly faster than CALIME, which incurs substantial overhead from its causal discovery phase. Despite the relative increase in cost, all ABLE variants complete in just a few seconds per explanation, making them practical for real-world use cases.

The increased runtime for ABLE is a direct trade-off for its substantially higher fidelity and stability. Notably, ABLE\_FGSM is highly query-efficient, requiring only 300 queries to target model (150 neighborhood points, yielding 300 training points for the local model) compared to the 5000 points typically used by baselines like LIME and GLIME. This efficiency, combined with superior performance, makes ABLE—particularly ABLE\_FGSM—a favorable choice in scenarios where querying the target model is expensive or high fidelity is critical.

\subsection{Discussion}

The performance differences among ABLE's variants highlight the role of the underlying adversarial attack. Iterative white-box approaches like PGD and etDeepFool generally achieve the highest fidelity, likely by finding points closer to the true boundary using gradients. The single-step FGSM, while faster, sometimes results in slightly lower fidelity. The black-box HSJ variant offers comparable stability and fidelity without requiring model internals, making it suitable for scenarios with limited access to the classifier. This suggests that even a query-based approximation of the boundary is more effective than unguided random sampling. The results also suggest that for some complex models like TabNet, finding minimal perturbations can be challenging, which may affect performance. This points to future work in developing even more robust adversarial search techniques tailored for local explanation.

\section{Conclusion and Future Works}

In this paper, we introduced our approach, ABLE, that constructs a local model to explain a model prediction by directly targeting the local decision boundary. By generating pairs of adversarial data points—one that minimally perturbs the test instance to flip the model’s prediction and another that reverts it back—we effectively bracket the local decision boundary of the target model for the test instance. Our experimental results demonstrate that ABLE not only mitigates the instability issues inherent in techniques like LIME but also enhances the fidelity of local models by anchoring them around the local decision boundary for the test instance. 

Future work includes evaluating ABLE on broader model types and data domains, such as images and text. When applied to image data, our approach must take into account the perceptual properties and spatial structure of the images. Applying perceptually motivated distance metrics like LPIPS \cite{zhang2018unreasonable} or SSIM \cite{wang2004image} during adversarial perturbations can help preserve spatial coherence.

\section*{Acknowledgment}
This work is supported by a research grant (70NANB21H092) from Information Technology Lab of National Institute of Standards and Technology (NIST).

Disclaimer: Certain equipment, instruments, software, or materials are identified in this paper in order to specify the experimental procedure adequately.  Such identification is not intended to imply recommendation or endorsement of any product or service by NIST, nor is it intended to imply that the materials or equipment identified are necessarily the best available for the purpose.

\bibliographystyle{ACM-Reference-Format}
\bibliography{KDD}

\end{document}